\pdfoutput=1

\documentclass[11pt]{article}

\usepackage[]{acl}

\usepackage{times}
\usepackage{latexsym}

\usepackage[T1]{fontenc}

\usepackage[utf8]{inputenc}

\usepackage{microtype}

\usepackage{inconsolata}

\usepackage{graphicx}
\usepackage{booktabs}
\usepackage{enumitem}
\usepackage{amsfonts}
\usepackage[most]{tcolorbox}
\usepackage{caption}
\usepackage{subcaption}

\usepackage{hyperref}
\usepackage{url}

\usepackage{xcolor}
\usepackage{xspace}

\usepackage{makecell}

\newcommand{\myparagraph}[1]{\paragraph{#1}}

\title{Interpreting Style Representations via Style-Eliciting Prompts}

\newcommand{\camerareadytext}[1]{#1}

\author{Junghwan Kim \and David Jurgens \\
University of Michigan\\
\texttt{\{kimjhj,jurgens\}@umich.edu}
}

\begin{document}
\maketitle
\begin{abstract}
Style representation learning is a powerful tool for authorship analysis and modeling writing style, yet the latent nature of learned representations makes them difficult to interpret.
Recent work has attempted to explain these representations by generating natural language descriptions with large language models (LLMs) conditioned on input text.
However, such descriptions are often prone to the LLM's biases and hallucinations, and they lack an explicit objective and practical utility.
In this work, we propose a novel framework for interpreting style representations through style-eliciting prompts: natural language instructions designed to steer LLMs to generate text that reflects specific stylistic attributes.
We curate 1{,}010 distinct style features spanning 26 stylistic categories and construct a dataset by prompting an LLM to generate text conditioned on these features.
Using this data, we train a decoder to generate a style prompt from the style representation of the generated text.
We evaluate our approach on three tasks:
(1) recovering original style prompts from generated text,
(2) generating text in the same style using the recovered prompts,
and (3) steering LLM outputs to match the style of human-written texts.
Experiments demonstrate that our method consistently outperforms strong baselines \camerareadytext{that directly prompt LLMs with target text, achieving superior performance in both style description and style imitation}.
These results highlight that style-eliciting prompts can provide a practical and interpretable interface to stylistic information encoded in style representations.
\end{abstract}

\section{Introduction}
Writing style is a core dimension of natural language, influencing how messages are interpreted, remembered, and disseminated across various contexts~\cite{kelly2003perceptions,boghrati2023style}.
To model stylistic variation computationally, recent work has developed \textit{style representations}---vector embeddings designed to encode stylistic properties~\cite{riverasoto2021luar,wegmann2022cav,patel2025styledistance}.
These representations have proven effective for modeling and comparing writing styles~\cite{neelakanteswara2024rags,soto2024fewshot,horvitz2024paraguide,horvitz2024tinystyler}.
Nevertheless, their latent nature obscures which stylistic attributes they encode, restricting how users can interact with them for controlled text generation.

One intuitive approach to creating interpretable style representations is to ask LLMs to describe the input text in natural language~\cite{patel2023lisa,alshomary2025latent}.
Yet, free-form LLM-generated descriptions often overlook important stylistic nuances and can be influenced by model-specific biases or hallucinations \citep{ramnath2025cave}.
More importantly, such descriptions are primarily explanatory rather than functional: it is not obvious how they can be reliably used to reproduce, manipulate, or transfer writing style.

In this work, we adopt a complementary perspective that emphasizes control as an explanation, building on LLMs' well-established capability to follow stylistic instructions when generating text~\cite{reif2022recipe}.
We introduce a framework that interprets neural style representations as human-language style prompts---explicit natural language instructions that specify stylistic constraints and can be directly followed by LLMs.
Our approach constructs a supervised learning setup in which texts are first generated from known style prompts, after which a decoder is trained to recover those prompts from the style representations of the generated texts.
By grounding interpretation in prompts that are directly usable for generation, this formulation yields an operational interface for stylistic control, supporting applications such as creative writing, personalized messaging, and persona simulation~\cite{mou2020stylizedtextgeneration}.

To support this approach, we build a large-scale synthetic dataset comprising 1{,}010 distinct style features organized across 26 stylistic dimensions, including lexical choice, syntactic structure, tone, and rhetorical strategy.
Using these prompts, we generate 1.8M stylized responses with an LLM, forming paired examples of text and explicit stylistic instructions.
This dataset provides fine-grained supervision for learning to decode style representations and enables systematic evaluation of both style characterization and controllability.

We demonstrate the benefits of our framework across three tasks:
(1) inferring original style prompts from LLM-generated text,
(2) producing stylistically similar outputs using inferred style prompts,
and (3) steering LLMs to emulate the style of non-synthetic, human-authored texts.
Across all evaluations, our method consistently surpasses baselines that rely on directly prompting LLMs with target text alone.
Our method achieves substantial gains in style prompt recovery (76.0\% ROUGE-1, 21.7\% LaBSE, and 42.8\% LLM-as-judge improvements) and stronger style alignment for style control (12.9\% and 26.1\% L2 improvement for LLM-generated and human-written references, respectively).
Together, these results demonstrate that decoding style representations into actionable prompts provides an effective pathway for both analyzing and manipulating writing style.

Our contributions are summarized as follows:
(1) We introduce a decoder-based framework that translates latent style representations into natural language style prompts, enabling interpretable and controllable use of stylistic information.
(2) We introduce a new, large-scale synthetic dataset of 1.8M stylized texts paired with diverse, compositional style prompts spanning 26 stylistic dimensions.
(3) We show that our method substantially outperforms existing baselines on both style prompt recovery and style control tasks.
(4) We release our dataset\footnote{\url{https://huggingface.co/datasets/Blablablab/style-to-text}} and code\footnote{\url{https://github.com/junghwanjkim/style-decoding}} to facilitate future research on interpretable and controllable writing style modeling.

\section{Related Work}
\label{sec:related_work}

Our work connects several research areas, including style representation interpretation, style description, style transfer, and prompt discovery.

\myparagraph{Style Representation.}
Originally developed for authorship verification, style representation models are trained to embed texts from the same author close together while separating those from different authors~\cite{riverasoto2021luar,wegmann2022cav,patel2025styledistance}. These dense vector representations have resulted in significant performance gains for many tasks like authorship attribution over older, more interpretable methods from stylometry \citep[cf.][]{rangel2019pan,stamatatos2022overview,bevendorff2024overview}.
Although their training objective focuses on author identity, the resulting representations capture rich stylistic features beyond author-specific ones.
For instance, controlled stylistic perturbations induce consistent linear shifts~\cite{zhu2021idiosyncratic} or clusters~\cite{wegmann2022cav} in the style representation space, indicating their sensitivity to stylistic changes.
Moreover, \citet{wang2023luar} verifies that various non-author-specific styles in the CDS dataset~\cite{krishna2020cdsdataset} can be successfully predicted from these representations.

\myparagraph{Style Representation Interpretation.}
Downstream applications of style representations---such as authorship analysis and writing style modeling---often require interpretability and transparency~\cite{tiersma2002forensic,biber2019register}.
To address this need, prior work has attempted to interpret style representations by aligning individual embedding dimensions with stylistic features~\cite{patel2023lisa} or by interpolating between representative examples in a latent style space~\cite{alshomary2025latent}.
However, in both cases, the resulting style descriptions are generated by prompting LLMs directly with the input text.
As a result, these descriptions may not faithfully reflect the stylistic information encoded in the style representations, nor are they tied to a concrete, reusable objective for generation or control.
In contrast, our framework defines explicit ground-truth style prompts that capture the stylistic intent injected into the style representations and can be used to steer subsequent text generation, enabling effective training supervision and well-defined evaluation.

\myparagraph{Style Description.}
Recent studies~\cite{hung2023verification,huang2024identification,hu2024instructav,ramnath2025cave} have explored generating style descriptions as intermediate steps for LLM-based authorship verification~\cite{stamatatos2016authorship,tyo2022state}, which aims to determine whether two documents share an author.
Other work~\cite{patel2024styll,yang2025registeranalysis} has used style descriptions to support LLM-driven style transfer~\cite{jin2022deeptstsurvey,hu2022tstreview,mukherjee2024tstsurvey}, where new text is generated with the content of one input and the style of another.
However, as \citet{ramnath2025cave} notes, such descriptions often fail to capture the full range of stylistic variation and are susceptible to biases and hallucinations inherited from the underlying LLM.
Our work addresses these limitations by treating style prompts as ground-truth descriptions, yielding more grounded, goal-directed, and operational characterizations of style.

\myparagraph{Style Transfer.}
The text style transfer literature has long investigated leveraging stylistic information represented as style vectors~\cite{hu2017toward,shen2017styletransfer,prabhumoye2018styletransfer,xu2020variational,shen2020educating}.
Recent work~\cite{horvitz2024paraguide,horvitz2024tinystyler} successfully trains text style transfer models that rely on pretrained style representations as input.
While our setting may appear similar due to its focus on stylistic control, our work differs fundamentally in both goal and formulation.
Rather than paraphrasing a given text to match a target style, our objective is to explicitly describe the target style and use this description to generate new text in that style.
Due to this difference, unlike in style transfer, where meaning preservation is often a central requirement, our setting does not involve preserving the original content; instead, we focus on evaluating the faithfulness of the style description and its utility for stylistic change.

\myparagraph{Prompt Discovery.}
The problem of discovering prompts that elicit specific behaviors from LLMs has attracted growing interest, particularly for uncovering harmful or undesirable behaviors~\cite{perez2022redteaming,liu2024autodan,hong2024curiositydriven}.
More recent work extends this paradigm to prompt search for arbitrary, user-defined objectives~\cite{li2025eliciting}.
Our study addresses a related but more fine-grained challenge: discovering prompts that induce specific writing styles---such as tone, sentence structure, or rhetorical form---which has not been explored in prior prompt discovery work.
General prompt discovery methods typically aim to find prompts that cause LLMs to generate a target text, often relying on reinforcement learning to explore the prompt space.
In contrast, our method leverages synthetic supervision to train a decoder efficiently in a fully supervised manner.

\section{Problem Formulation}
\label{sec:problem}
We study the interpretation of style representations by mapping them into natural-language descriptions.
Specifically, we formulate style representation interpretation as the task of decoding style representations into style prompts that can be used to steer LLMs to produce text in the style implicitly described by the representation.
Given a style representation of an input text, the objective is to infer a natural-language prompt that (1) induces LLM generations whose style is consistent with the stylistic intent encoded in the representation and (2) provides an interpretable description that meaningfully characterizes the style.
This formulation is motivated by three considerations: (i) such prompts are easily understandable by humans, (ii) they enable principled evaluation by assessing whether generated text exhibits the target style, and (iii) they are directly usable for downstream control of LLM generation.

Formally, let $\mathtt{S}$ denote a Style Representation Model (SRM) that maps an input text $x$ to a latent style vector $\mathbf{x}$.
Our goal is to learn a decoder $\mathtt{D}$ that maps $\mathbf{x}$ to a style prompt $s$ which both characterizes the stylistic features encoded in $\mathbf{x}$ and steers a subsequent generation $y$ to exhibit a style similar to that of $x$.
This objective can be expressed as
\begin{equation}
\label{eq:objective}
    {\arg\min}_{\mathtt{D}} \quad \ell(\mathbf{x}, \mathbf{y}),
\end{equation}
where $\mathbf{y} = \mathtt{S}(y)$ denotes the style vector of the generated text $y$, and $\ell$ is a vector distance (e.g., L2).
Appendix Figure~\ref{fig:problem_formulation_1} illustrates this formulation.

Directly optimizing Equation~\ref{eq:objective} is infeasible due to the vast and discrete nature of the prompt space.
To address this challenge, we recast the problem as supervised learning using synthetic data.
Specifically, we generate synthetic pairs $(x, s)$ consisting of a stylized text $x$ and its corresponding style prompt $s$,
and train the decoder $\mathtt{D}$ to map the style vector $\mathbf{x} = \mathtt{S}(x)$ to the ground-truth prompt $s$:
\begin{equation}
\label{eq:surrogate}
    {\arg\min}_{\mathtt{D}} \quad \mathcal{L}(\tilde{s}, s),
\end{equation}
where $\tilde{s} = \mathtt{D}(\mathbf{x})$ denotes the decoded style prompt, and $\mathcal{L}$ is the token-level cross-entropy loss.
This surrogate objective is depicted in Appendix Figure~\ref{fig:problem_formulation_2}.
If $\mathtt{D}$ perfectly recovers the ground-truth prompt $s$, then both input text $x$ and the generated text $y$ in the original formulation are produced using the same prompt $s$.
Consequently, their style vectors $\mathbf{x}$ and $\mathbf{y}$ are closely aligned, thereby minimizing the objective in Equation~\ref{eq:objective}.

To construct synthetic training pairs $(x, s)$, one possible approach is to use LLMs to describe the style of existing texts.
However, such descriptions may overlook important stylistic nuances and can be affected by model-specific biases or hallucinations.
Moreover, many texts may not exhibit sufficiently salient stylistic features to support reliable description.
Instead, we begin with diverse style prompts and generate stylized texts using LLMs.
Because LLMs demonstrate strong instruction-following capabilities for stylistic control, these prompts reliably induce the salient stylistic features expressed in the generated texts.
Furthermore, since the prompts are used directly to generate the texts, they are guaranteed to be operational and can steer the style of LLM generations.

\begin{figure*}[t]
    \centering
    \includegraphics[width=\textwidth]{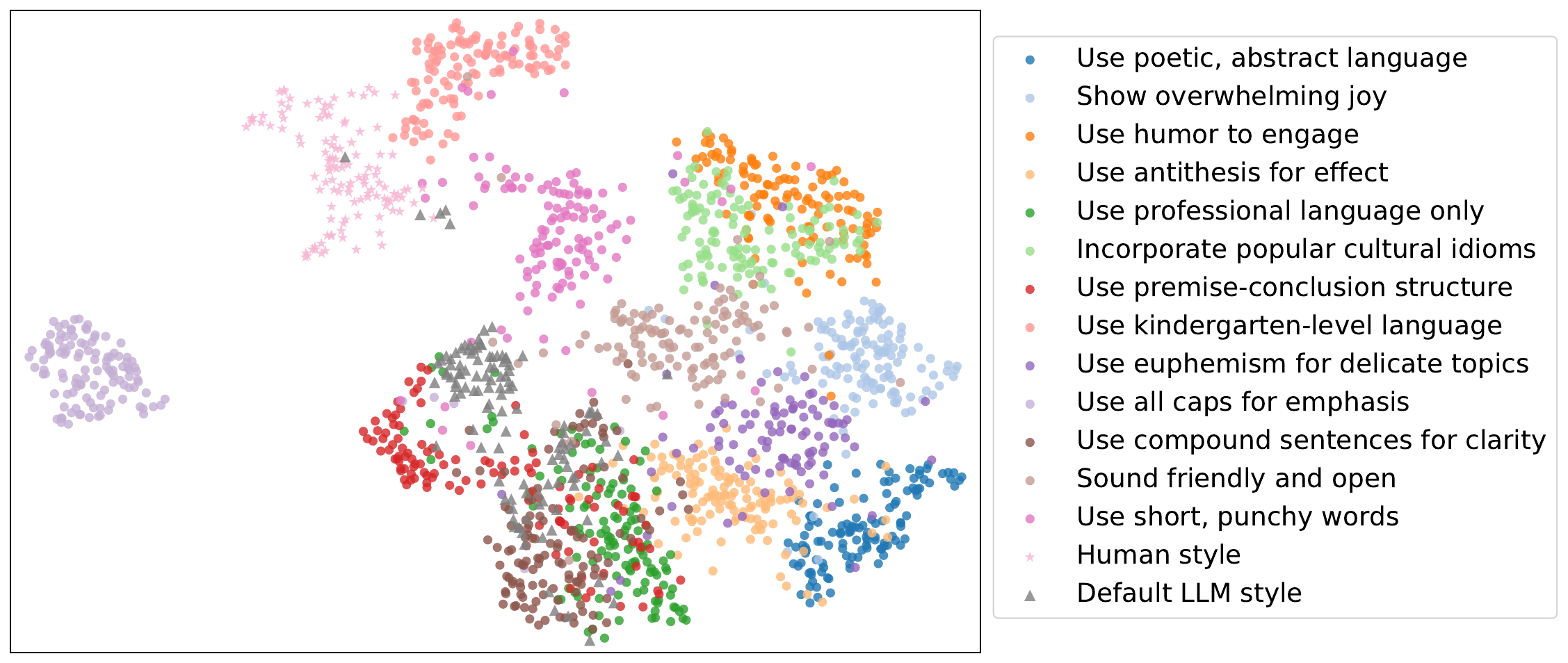}
    \caption{We visualize the style representations of a subset of our final dataset using t-SNE.
    Generations conditioned on different style prompts form distinct clusters, with stylistically similar clusters appearing closer together, illustrating that style representations admit style prompt recovery.}
    \label{fig:embedding}
\end{figure*}

\section{Dataset Construction}
\label{sec:data}
We construct a large-scale synthetic dataset of stylistically diverse texts paired with ground-truth style prompts that explicitly characterize their writing style.
To this end, we generate LLM outputs conditioned on a wide range of style prompts and real-world questions drawn from public Question-Answering (QA) platforms, ensuring content diversity and facilitating direct comparison with human-written answers.
The dataset construction process consists of three stages: (1) generating a large inventory of concrete style features, (2) curating diverse real-world questions, and (3) generating stylized responses conditioned jointly on style prompts and questions.
We describe each stage below.

\myparagraph{Style Generation.}
A central component of our dataset is a diverse set of modular style features that capture a broad spectrum of stylistic variation.
These features can be composed into prompts to control multiple stylistic attributes.

We generate the style feature set using GPT-4o with a hierarchical prompting strategy.
Directly prompting for a flat list of features leads to substantial redundancy and uneven coverage.
Instead, we first define broad stylistic categories and then populate each category with fine-grained features.

Our category design draws on prior work and is further expanded using LLM assistance.
We begin with 12 seed categories derived from existing literature~\cite{fisher2024styleremix,patel2025styledistance,ramnath2025cave} and expand this set to 26 categories using GPT-4o.
These categories span a wide range of stylistic dimensions, including ``Sentence Structure and Syntax,'' ``Word and Expression Usage,'' ``Tone and Mood,'' and ``Readability Level.''
The complete list of categories is shown in Table~\ref{tab:data_style_category}.

Each category is populated with style features through LLM generation, followed by manual curation.
For each category, GPT-4o is prompted to generate 40 concrete, modular, and semantically distinct style features.
We then manually filter redundant or overly similar entries both within and across categories.
The resulting feature set contains 1{,}010 curated style features spanning diverse categories.
Examples of these features are shown in Figure~\ref{fig:embedding}, with additional details and curation prompts provided in Appendix~\ref{subsec:appendix_data_style}.

\myparagraph{Question Curation.}
We ground stylized generation in a QA setting to enable diverse outputs under the same style prompt.
This setup supports open-ended yet content-grounded responses, allowing models to vary content while adhering to a specified style.
By pairing a single style prompt with different questions, LLMs can generate stylistically consistent but semantically distinct outputs.

We curate 300{,}000 real-world questions from 3 publicly available QA platforms: Reddit, StackExchange, and Yahoo Answers.
For each platform, we select 10 topics\footnote{Topics correspond to subreddits for Reddit and sites for StackExchange.} and sample 10{,}000, 5{,}000, and 15{,}000 questions per topic from Reddit, StackExchange, and Yahoo Answers, respectively.
The resulting questions span a broad range of domains, including technical problem solving, personal advice, and opinion-based discussion.
The full list of topics is provided in Table~\ref{tab:data_question_topic}, with further details in Appendix~\ref{subsec:appendix_data_question}.

To support evaluation against human-written references, we also collect corresponding human answers for the curated questions.
We first filter out questions with no human responses, then randomly select one answer per remaining question.

\myparagraph{LLM Generation.}
We generate stylized responses by prompting LLMs with both a style prompt and a question.
Each style prompt is constructed by randomly sampling between 1 and 10 style features from the curated feature set and concatenating them into a single instruction.
To avoid conflicting stylistic constraints, we ensure that features within a prompt are drawn from distinct style categories.
The sampling distribution favors prompts with fewer features, enabling the decoder to first learn individual stylistic attributes before generalizing to compositional styles.

Each question is paired with two distinct style prompts, yielding two stylized responses per question.
To mitigate model-specific stylistic biases, we generate responses using three different LLMs: Phi-4, Qwen2.5-14B, and OLMo-2-13B.
Additional details on prompting and generation settings are provided in Appendix~\ref{subsec:appendix_data_generation}.

\myparagraph{Final Dataset.}
Each dataset entry consists of a question, a style prompt, and a stylized response.
Together, these examples provide explicit supervision for training the decoder to recover style prompts from style representations.
The final dataset comprises 1.8M LLM-generated responses with 434{,}535 unique style prompts, covering a wide range of stylistic variations and content domains.
In addition, the dataset includes 300K human-written responses to support comparison with human-authored style.

Figure~\ref{fig:embedding} visualizes a subset of our dataset in the style representation space.
Distinct clusters emerge, each corresponding to a different style prompt, with semantically similar prompts forming nearby clusters.
For instance, the prompts ``Use professional language only'' and ``Use compound sentences for clarity'' produce overlapping clusters.
Notably, the default LLM style lies close to professional language, while human-written responses cluster near short, punchy wording.
These patterns validate that the geometry of the style representation space captures stylistic similarity and variation, providing evidence that the representation space admits a decoding into interpretable style prompts.


\section{Proposed Method}
\label{sec:method}
We now describe the architecture and training of our proposed style decoder.
As formulated in Section~\ref{sec:problem}, the goal is to learn a decoder model $\mathtt{D}$ that maps a style vector $\mathbf{x}$ to a corresponding natural-language style prompt $s$.

Our decoder architecture is inspired by continuous prompt tuning, which has been shown to be effective for conditioning frozen LLMs in both NLP~\cite{li2021prefixtuning,lester2021prompttuning} and vision–language modeling~\cite{tsimpoukelli2021frozen}.
In this framework, continuous prompts encode input- or task-specific information that guides downstream generation.
We extend this paradigm by encoding stylistic information captured in style vectors into continuous prompts.

\begin{figure}[t]
    \centering
    \includegraphics[width=\columnwidth]{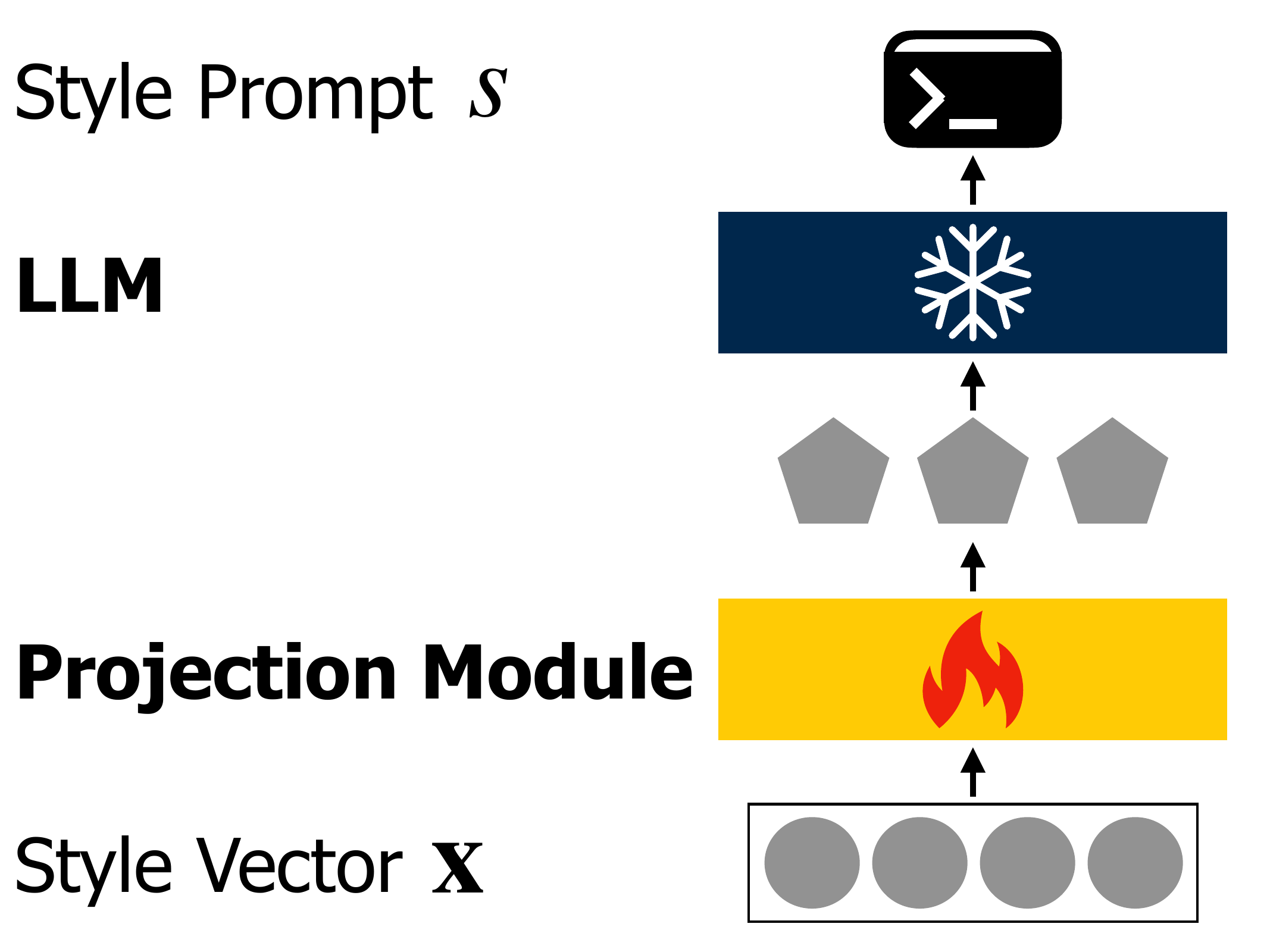}
    \caption{Our decoder $\mathtt{D}$ consists of a frozen LLM and a trainable projection module.
    The projection module maps the input style vector into the token embedding space of the LLM.
    Then, LLM generates a style prompt conditioned on the projected style vector.}
    \label{fig:model_architecture}
\end{figure}

The decoder $\mathtt{D}$ consists of two components:
(1) a frozen LLM that interprets the stylistic signal encoded in the style vector, and
(2) a trainable projection module that maps the style vector into the token embedding space of the LLM.
Because the projection module is the only trainable component, this module must be equipped with sufficient capacity to express a wide range of stylistic signals.
Concretely, the projection is implemented as a three-layer feedforward network with GeLU activations~\cite{hendrycks2023gelu}, producing 20 token embeddings, a choice that we found to work well in preliminary experiments.
To further contextualize decoding, we prepend a natural-language instruction that explicitly specifies the style description task alongside the projected style vector, enabling the LLM to more effectively leverage its instruction-following capabilities.
The instruction prompt is provided in Appendix~\ref{subsec:appendix_method_prompt}.
The overall architecture is illustrated in Figure~\ref{fig:model_architecture}.

\myparagraph{Implementation Details.}
For the style representation model $\mathtt{S}$, we use Mistral-Nemo-Instruct-2407, trained via contrastive learning on author-labeled data following the state-of-the-art approach of \citet{fincke2024separating}.
The frozen LLM used in the decoder $\mathtt{D}$ is Ministral-8B-Instruct.

We train the decoder $\mathtt{D}$ using the surrogate objective defined in Equation~\ref{eq:surrogate}.
The dataset is split into training, validation, and test sets with a ratio of 8:1:1.
Training is performed for 5 epochs with a learning rate of 5e-5 and a batch size of 32.
The best checkpoint is selected based on validation loss, and all reported results in this paper are obtained on the test set\footnote{180K LLM responses for Sections~\ref{sec:exp1} and \ref{sec:exp2} and 60K human responses for Section~\ref{sec:exp3}}.
Additional training details are provided in Appendix~\ref{subsec:appendix_method_training}.


\begin{figure*}[t!]
\centering
\includegraphics[width=\textwidth]{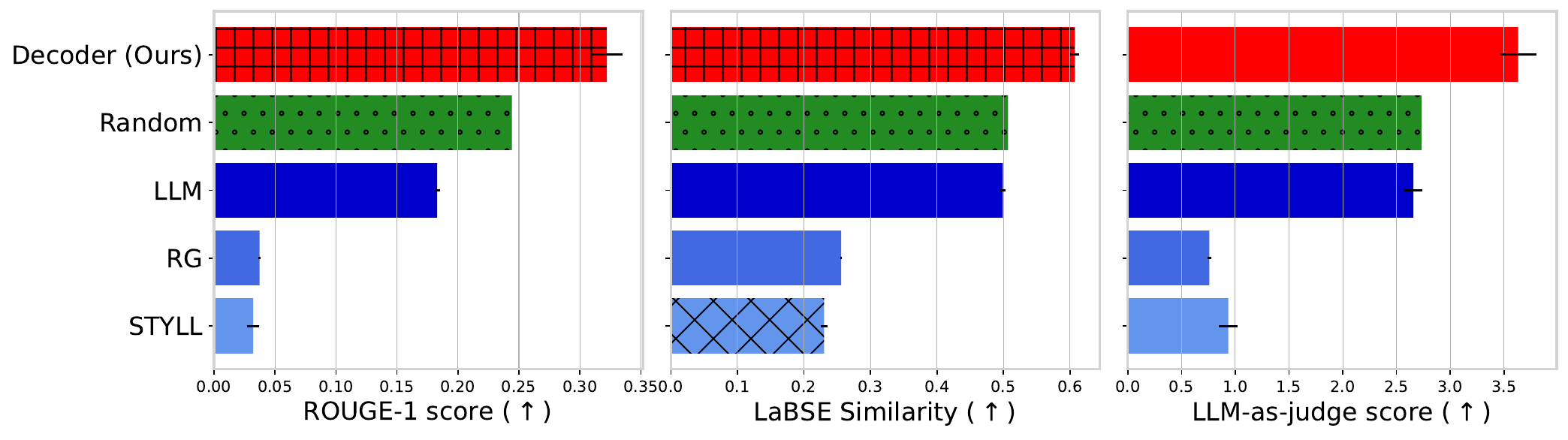}
\caption{The style descriptions generated by our decoder match the ground truth style prompts better than the style descriptions directly generated by LLM or randomly selected style prompts in our dataset.
\camerareadytext{The match is consistently demonstrated in all three metrics: ROUGE-1 score (token matching), LaBSE similarity (semantic embedding) and LLM-as-judge score (LLM judgment).}}
\label{fig:exp1_results}
\end{figure*}

\section{Recovering the Style Prompt}
\label{sec:exp1}
We first evaluate prompt recovery, the task of reconstructing the ground-truth style prompt from the style representation of a text generated by an LLM conditioned on that prompt.
This task directly corresponds to the surrogate training objective defined in Equation~\ref{eq:surrogate}.
The goal of this experiment is to assess whether the proposed decoder can effectively recover the original stylistic instruction encoded in the style representation---i.e., creating an interpretable representation of the style.

\subsection{Experiment Setup}

Given a style prompt $s$ and the corresponding stylized response $x$, our decoder maps the style representation $\mathbf{x}$ of $x$ to the decoded style prompt $\tilde{s}$.
We measure how well $\tilde{s}$ matches the ground truth $s$.

\myparagraph{Baselines.}
We compare our approach against four baselines.
The first three are LLM-based description baselines, in which an LLM is prompted to describe the writing style of a given text in natural language.
Specifically, we adapt prompts from two prior works, STYLL~\cite{patel2024styll} and RG~\cite{yang2025registeranalysis}, that generate style descriptions for style transfer, with only minor modifications necessary for our setting.
Because these prompts were originally designed for style transfer, we additionally introduce a custom prompt baseline tailored to prompt recovery.
To minimize performance differences arising from format mismatch, the custom baseline instructs the LLM to match the length and sentence structure of the desired output.
All description baselines use Ministral-8B-Instruct, the same LLM employed in our decoder, ensuring a fair comparison.
The full instruction prompts are provided in Appendix~\ref{subsec:appendix_exp1_baseline}.
The final baseline is random sampling, where a style prompt is drawn uniformly from the same distribution used during dataset construction.
This baseline serves as a naive lower bound on prompt recovery performance.

There is no straightforward off-the-shelf baseline that consumes dense style vectors.
We leave the comparison of different instantiations of our framework: mapping style vector to style prompt as future work.

\myparagraph{Metrics.}
We evaluate prompt recovery using three complementary metrics that capture both lexical and semantic similarity.
ROUGE-1 measures unigram overlap between the decoded prompt and the ground-truth prompt.
While this metric closely aligns with the training objective, it may underestimate similarity when stylistically equivalent prompts differ in surface wording.
To account for this limitation, we additionally report semantic metrics.
LaBSE similarity computes cosine similarity between sentence embeddings of the decoded and ground-truth prompts, while LLM-as-judge employs an external LLM (Qwen3-14B) to rate prompt similarity on a 0–10 scale.
The evaluation prompt is provided in Appendix~\ref{subsec:appendix_exp1_evaluation}.
Importantly, the evaluation model is entirely disjoint from those in dataset generation, decoder training, and baseline methods, ensuring an unbiased assessment.

\subsection{Results}
As shown in Figure~\ref{fig:exp1_results}, our decoder consistently outperforms all baselines across all three evaluation metrics by a significant margin.
This result indicates that the decoder is highly effective at recovering the original style prompt in both lexical agreement and semantic similarity.

All LLM-based descriptions consistently perform worse than the random baseline.
LLMs fail to generate style descriptions that capture the underlying stylistic intent expressed in the style prompt.
This is consistent with the observation in \citet{berglund2024reversalcurse} that even though an LLM generates B from A, it cannot infer that A could have generated B.
Our custom baseline seems to benefit from output format and length information, but it still performs worse than our decoder.

Overall, these results demonstrate that decoding directly from style representations yields more accurate reconstructions of stylistic intent than post hoc LLM-based description methods.
Moreover, they provide strong evidence that the style representation space encodes rich, fine-grained stylistic information, and that our decoder can effectively translate these latent signals into interpretable natural-language prompts.

\begin{table*}[t!]
\centering
    \begin{tabular}{l}
    \textbf{Example Outputs} \\
    \hline
    \makecell[l]{1. The author uses unreliable memory style, uses formal vocabulary, uses interjections for surprise,\\ uses high noun density, uses repetition for focus, uses commas to set off interrupters,\\ uses a flowchart-like logic.} \\
    \makecell[l]{2. The author uses poetic descriptive style, uses anaphora in sentence beginnings.} \\
    \makecell[l]{3. The author uses comparative adjectives for comparison, uses minimal supporting explanation,\\ uses antithesis in sentence structure.}
    \end{tabular}
\caption{The example outputs of our decoder.}
\label{tab:exp1}
\end{table*}

Example output is shown in Table~\ref{tab:exp1}.
From our inspection of randomly sampled examples, we do not observe content leakage---i.e., the decoded style is not somehow leaking information about the content of the text that would make inferring its style easier. Instead, all output follows the format we used to train the decoder, making our evaluative comparisons meaningful.


\section{Controlling the Writing Style}
\label{sec:exp2}
We next examine whether decoded style prompts can functionally control the style of LLM outputs.
In contrast to prompt reconstruction, this setting does not require recovering the original instruction verbatim.
Instead, success is determined by whether the decoded prompt induces generations with the same stylistic effect as those produced under the ground-truth prompt.
It therefore serves as a test of whether training with the surrogate objective in Equation~\ref{eq:surrogate} yields prompts that generalize to effective style control in downstream generation.

\subsection{Experiment Setup}

Given a style vector $\mathbf{x}$ and the style prompt $\tilde{s}$ decoded from $\mathbf{x}$, we generate a new text $y$ conditioned on $\tilde{s}$.
We measure how close the style representation $\mathbf{y}$ of $y$ is to the original $\mathbf{x}$.
To avoid information leakage, $x$ and $y$ are conditioned on different question contexts.

\myparagraph{Baselines.}
We compare our approach against five baselines: four LLM-based style-imitation methods and one explicit style-transfer approach.
For the LLM-based imitation baselines, we adapt prompting strategies from three prior works~\cite{bhandarkar2024emulate,wang2025imitate,jangra2025sptg}, which are designed to generate responses that mimic the writing style of a given target text.
We modify these prompts minimally to fit our QA setup.
In addition, we again introduce a custom LLM baseline tailored to our setting.
The full prompts are provided in Appendix~\ref{subsec:appendix_exp1_baseline}.
Both our method and the LLM-based imitation baselines use Ministral-8B-Instruct for generation.
As a representative style transfer baseline, we include TinyStyler~\cite{horvitz2024tinystyler}, a state-of-the-art method that applies style representations to transform text.
We use the official implementation, which uses its own style representation.
We first generate neutral responses without style conditioning and then apply TinyStyler to transfer them into the target style.

\myparagraph{Metrics.}
We assess style control by computing the L2 distance between the style representations of the original and the regenerated text.
Our evaluation uses the same style representation model that encodes inputs for the decoder, since the goal of our framework is to faithfully interpret the style representation space.
The decoded prompts should have a consistent effect on the target style representation space, rather than on obscure general style.

\begin{figure}[t!]
    \centering
    \includegraphics[width=\columnwidth]{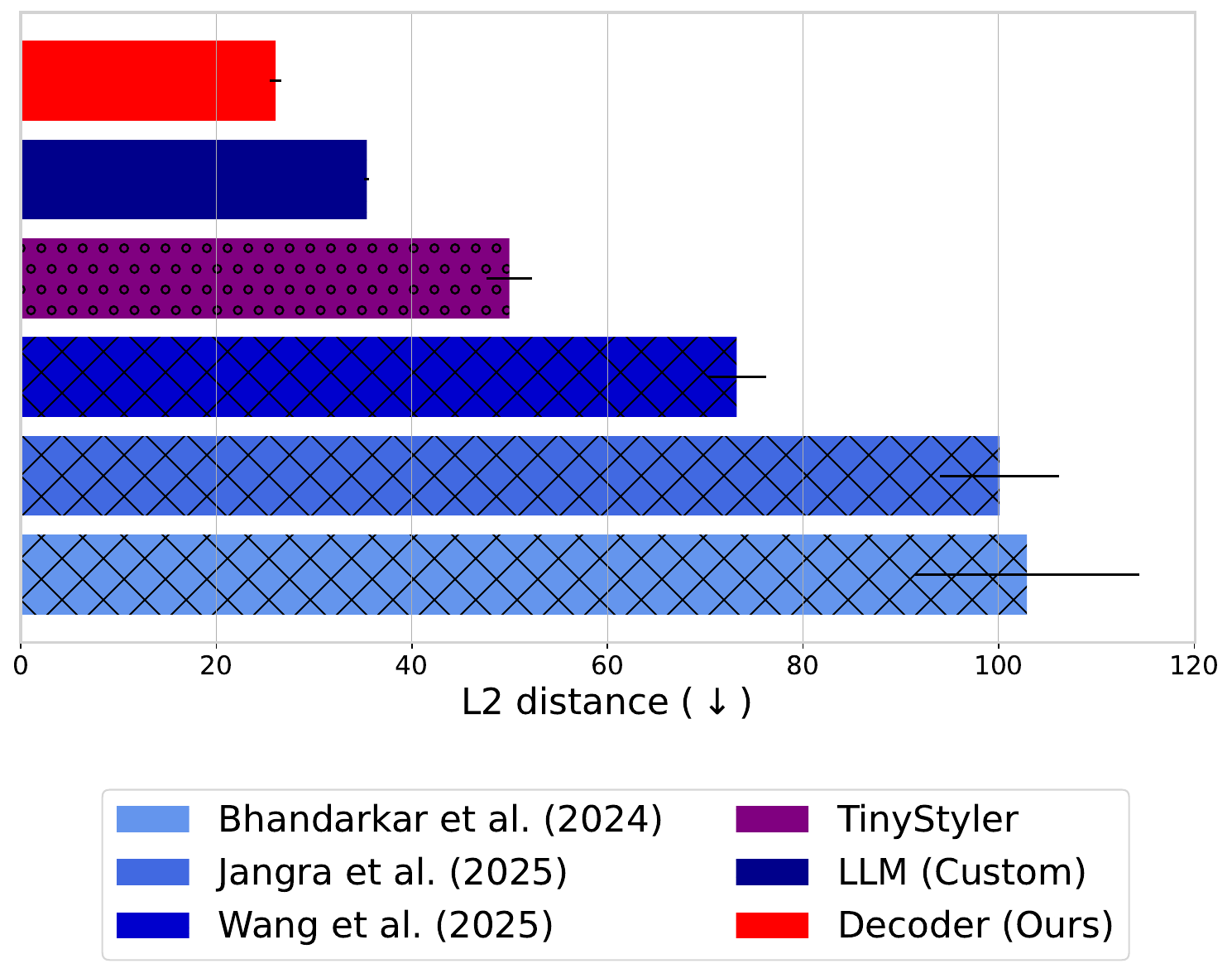}
    \caption{Our approach yields the smallest L2 distance in style representation space from the ground-truth stylized output, compared to baseline methods.}
    \label{fig:exp2_results}
\end{figure}

\subsection{Results}
Figure~\ref{fig:exp2_results} shows that our decoder outperforms all baselines by a significant margin.
This result indicates that the inferred prompts successfully translate latent stylistic information into actionable instructions that guide generation behavior.

%
The TinyStyler outperforms three adapted LLM-based style-imitation baselines, but underperforms our custom baseline.
One plausible limitation of the TinyStyler is architectural: TinyStyler relies on a smaller T5-based generator, whereas both our decoder and the LLM-based baselines use LLMs with billions of parameters.
In addition, TinyStyler employs the StyleEmbedding model~\cite{wegmann2022cav}, while our decoder leverages a substantially larger style representation model trained on a much broader, author-labeled corpus.

As an additional robustness test, we repeat the evaluation using two alternative style representations not used in our method, LUAR \cite{riverasoto2021luar} and StyleDistance \cite{patel2025styledistance}.
The results, provided in  Appendix~\ref{subsec:appendix_exp2_additional}, show consistent trends across all three style representations, with our approach able to generate new documents that are closer in the vector space, regardless of which representation is used. Together, these results demonstrating that the strong steering performance of our decoder is not specific to the style representation space used for training it.


\begin{figure}[t!]
    \centering
    \includegraphics[width=\columnwidth]{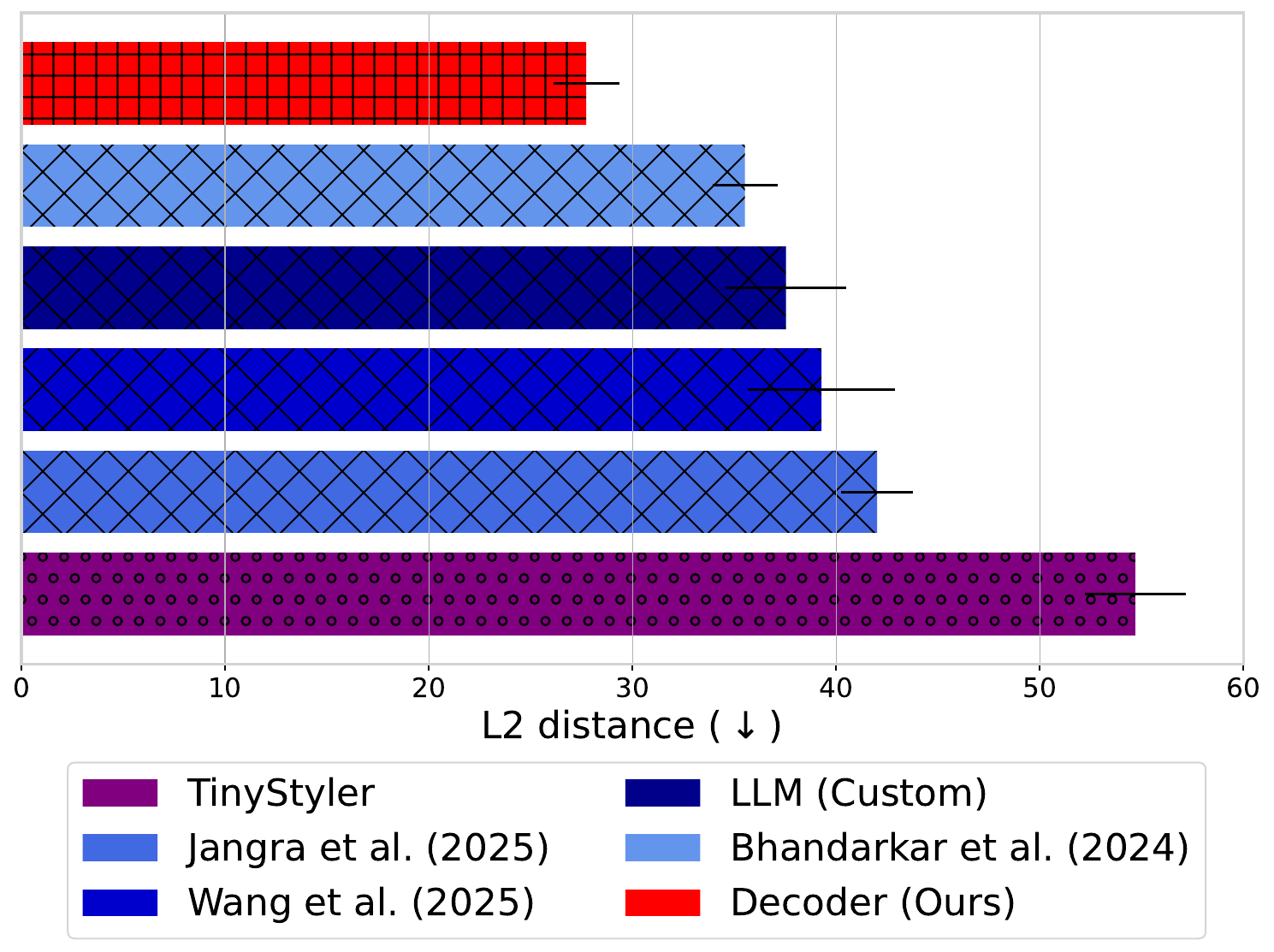}
    \caption{Our decoder outperforms baselines at generating style prompts that induce writing most similar to human-written text.}
    \label{fig:exp3_results}
\end{figure}

\section{Steering towards Human Style}
\label{sec:exp3}
Finally, we investigate whether the proposed decoder can steer LLM generation toward the writing style of human-authored text.
This experiment evaluates generalization to natural, non-synthetic language outside the decoder's training distribution, which constitutes a more realistic and challenging test of practical style control.
%

\subsection{Experiment Setup}
We are given a style vector $\mathbf{x}$ and the style prompt $\tilde{s}$ decoded from $\mathbf{x}$, as in Section~\ref{sec:exp2}.
However, the target text $x$ is now human-authored text rather than LLM-generated text, thereby falling outside the distribution of the synthetic training data.
Therefore, the decoder operates on the input distribution unseen during training, and there is no ground-truth style $s$ that guarantees the generation in the style of $x$.
This setting makes the task more challenging, but generalization to it is essential for the decoder to work in a more practical style control.
We use the same metric and baseline as in Section~\ref{sec:exp2}.

\subsection{Results}
As shown in Figure~\ref{fig:exp3_results}, generations guided by decoded style prompts achieve substantially smaller distances to human-authored references than those produced by any baseline method.
Because no ground-truth style prompts exist for human-authored texts, strong performance in this setting provides a challenging test of the proposed framework.
The results indicate that the decoder can effectively interpret input style vectors and map them to appropriate style prompts even when the inputs lie outside the distribution of the synthetic training data.
Moreover, the diversity of the curated style features enables generalization across a broad spectrum of stylistic variation, extending to naturally occurring human writing.
Taken together, the results offer compelling evidence for the robustness and transferability of the decoding framework.

We repeat our robustness tests by using two alternative style representations LUAR and StyleDistance, to measure whether the generated style-steered document is similar to the reference human-authored document's style. Results shown in Appendix~\ref{sec:appendix_exp3} demonstrate a consistent trend with our approach performing best for matching the style. Together, these results show that the style instructions from our decoder successfully steer style towards human texts.





\section{Conclusion}
\label{sec:conclusion}
We propose a novel framework for interpreting dense style representation vectors by decoding them into natural language descriptions of writing style.
To train a decoder in a supervised setting, our approach leverages synthetic data generated via LLM with human-readable, compositional style prompts.
Comprehensive evaluations demonstrate that the decoded prompts faithfully reflect the stylistic information encoded in style representations and can effectively guide LLM outputs to match a desired style.
\camerareadytext{Our method consistently outperforms baselines across multiple metrics, including lexical and semantic match in style prompt recovery and style similarity in style control.}

This work bridges the gap between latent neural representations and interpretable, controllable style attributes, offering a path toward more transparent and steerable NLP systems.
Future directions include extending the method to zero-shot or unsupervised settings, improving disentanglement between content and style, and applying the decoding framework to other latent attributes such as tone, persona, or rhetorical strategy.

\section*{Limitations}
Our study focuses on the English language.
Since LLMs and style representation models that our framework builds on are less available or less performant in non-English languages, our method may not generalize to such settings.
Moreover, writing styles in different languages can have different distributions and interactions.
We hope that our work inspires other researchers to generalize our framework to non-English languages.

Our dataset is restricted to the online question-answering domain.
While our decoder demonstrates strong performance in describing writing style within this context, it remains unclear how well it generalizes to other domains such as narrative writing, formal prose, or technical documentation.
Evaluating the method across a broader range of domains is an important direction for future work.

Additionally, our dataset is generated using three similarly-sized LLMs.
Although we observe consistent style description quality across these models, this setup does not fully capture the diversity of existing LLMs in terms of scale, architecture, and training data.
To mitigate this, we test our decoder on human-written text and observe promising results.
Nevertheless, caution is warranted when applying the method to text produced by models that differ significantly from those used during dataset construction, as performance may degrade in unseen generative regimes.

\section*{Ethical Considerations}
This work presents a framework for analyzing the writing style of a given text and inferring a style prompt capable of eliciting similar stylistic characteristics from an LLM.
While this technique can support beneficial applications---such as improved interpretability, stylistic control, and enhanced transparency in LLM-generated text---it also introduces potential risks.
For example, it could be misused to undermine author anonymity in sensitive contexts or to reverse-engineer proprietary or private prompt formulations.

We emphasize that our approach is designed for controlled settings with synthetic supervision and does not aim to de-anonymize authors or recover confidential prompts.
Nevertheless, we encourage future work to explore safeguards and use restrictions to mitigate misuse, particularly in high-stakes or privacy-sensitive applications.

\section*{Acknowledgments}

This research is supported in part by the Office of the Director of National Intelligence (ODNI), Intelligence Advanced Research Projects Activity (IARPA), via the HIATUS Program contract \#2022-22072200006. The views and conclusions contained herein are those of the authors and should not be interpreted as necessarily representing the official policies, either expressed or implied, of ODNI, IARPA, or the U.S. Government. The U.S. Government is authorized to reproduce and distribute reprints for governmental purposes notwithstanding any copyright annotation therein.

\bibliography{reference}

\newpage
\appendix

\setcounter{figure}{0}
\setcounter{table}{0}
\renewcommand*\thefigure{A\arabic{figure}}
\renewcommand*\thetable{A\arabic{table}}

\section{Artifact for Reproducibility}
Table~\ref{tab:llms} shows the models and their size used in our experiments.

\section{Supplementary Material for Section~\ref{sec:problem}}

\begin{figure}[t]
    \centering
    \begin{subfigure}[b]{\columnwidth}
        \centering
        \includegraphics[width=0.9\columnwidth]{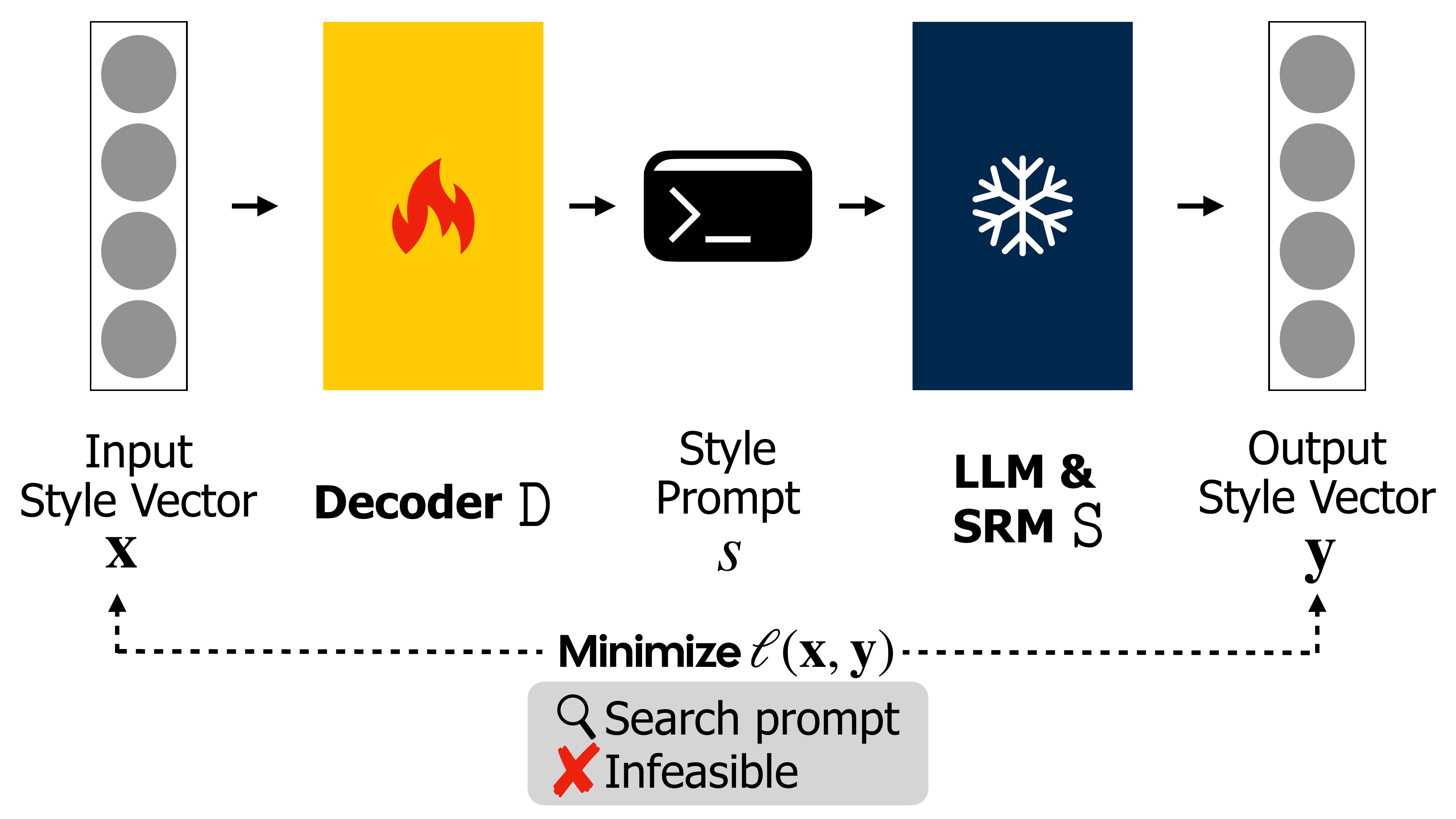}
        \caption{Prompt Search Formulation in Equation~\ref{eq:objective}.}
        \label{fig:problem_formulation_1}
    \end{subfigure}
    \\
    \begin{subfigure}[b]{\columnwidth}
        \centering
        \includegraphics[width=0.9\columnwidth]{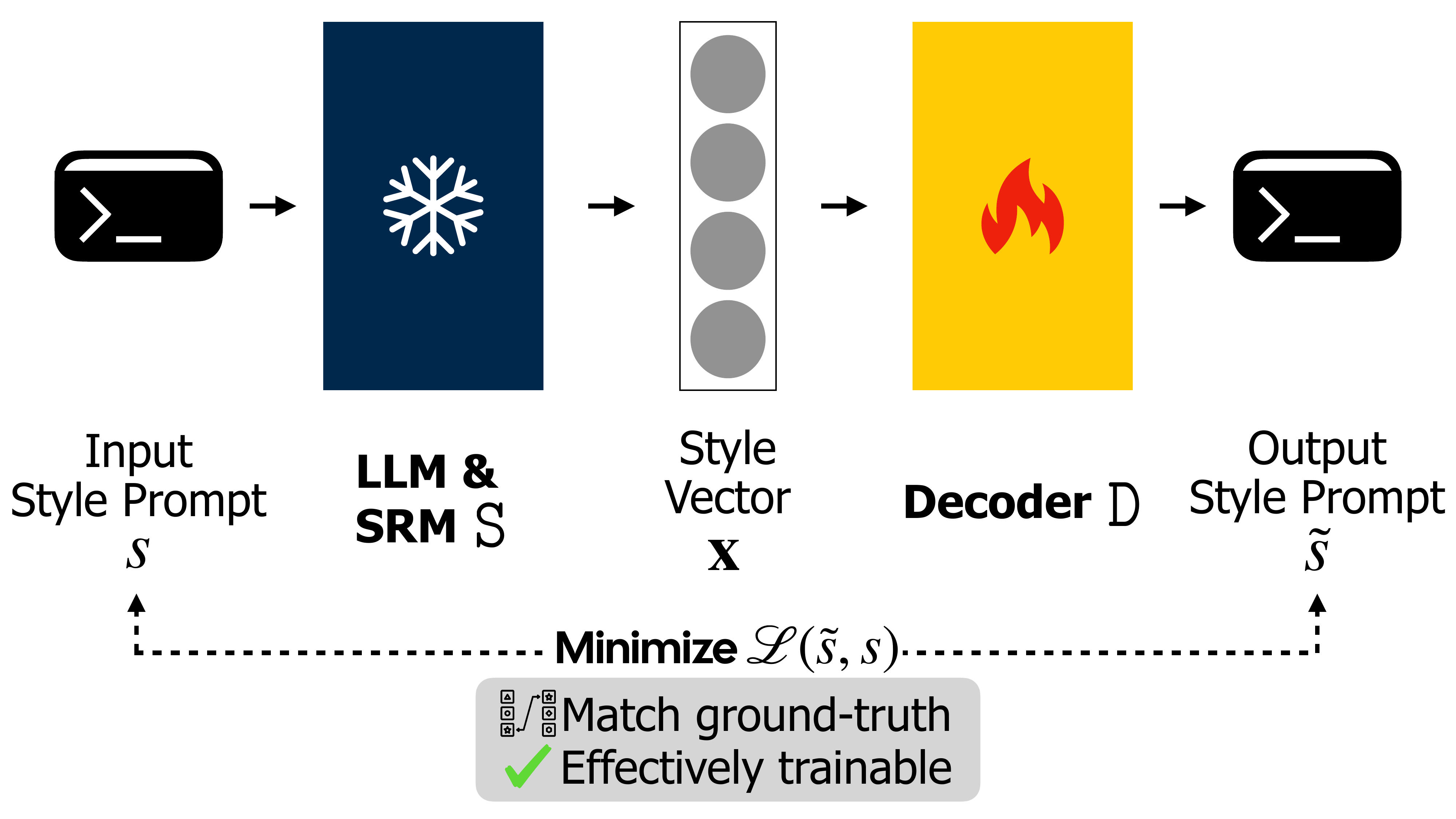}
        \caption{Prompt Recovery Formulation in Equation~\ref{eq:surrogate}.}
    \label{fig:problem_formulation_2}
    \end{subfigure}
    \caption{Our goal is to train a decoder $\mathtt{D}$ that recovers a style prompt $s$ capable of steering an LLM to generate text $y$ in the style of $x$ (Figure~\ref{fig:problem_formulation_1}).
    Because searching the vast prompt space is intractable, we reformulate the task as supervised prompt recovery: Given a style-response pair $(s, x)$, we match the decoded prompt $\tilde{s} = \mathtt{D}(\mathbf{x})$ to $s$ (Figure~\ref{fig:problem_formulation_2}).}
    \label{fig:problem_formulation}
\end{figure}

Figure \ref{fig:problem_formulation} illustrates the core problem being addressed.
Instead of the prompt search formulation in Figure~\ref{fig:problem_formulation_1} which is infeasible, we reformulate the task as the supervised prompt recovery (Figure~\ref{fig:problem_formulation_2}).

\section{Supplementary Material for Section~\ref{sec:data}}
\label{sec:appendix_data}
This section provides additional details on the dataset construction process.

\subsection{Style Generation}
\label{subsec:appendix_data_style}
We describe here the procedure used to obtain the style features.
First, we employ the \textbf{Category expansion prompt} below to obtain the style categories.
Table~\ref{tab:data_style_category} lists the stylistic categories in our dataset.
Next, we apply the \textbf{Feature list prompt} below, in which \texttt{[Style Category]} is replaced with the corresponding category name to produce fine-grained style features.

\myparagraph{Category expansion prompt.}
\texttt{\\
Extend the list of categories for writing styles. Each category should be distinct and orthogonal from the others. Provide only new categories.\\\\
``Writing Goal and Intent'', ``Tone and Mood'', ``Grade Level'', ``Sentence Structure and Syntax'', ``Function Word Usage'', ``Word and Expression Choice'', ``Punctuation Usage'', ``Special Character and Capitalization Usage'', ``Acronym and Abbreviation Usage'', ``Formality'', ``Social and Interpersonal'', ``Emotional Intensity'', ``Descriptive Density''
}

\myparagraph{Feature list prompt.}
\texttt{\\
Your task is to generate a list of 40 writing styles in the [Style Category] category. Answer in JSON format, where each item is a style.\\\\
Detailed Instructions:
Each style in the list should be
(1) distinct and orthogonal to the others. Similar styles with subtle nuance differences are considered distinct and, therefore, allowed.
(2) independent of the content of the text.
(3) concrete so that LLMs can implement it when instructed to.
(4) minimal so that it cannot be decomposed into multiple styles.
(5) a short sentence (3-5 words) in the imperative form that instructs writers to implement the style.
}

\subsection{Question Curation}
\label{subsec:appendix_data_question}
To ensure diversity across content domains, we sample questions stratified by topic.
For each platform, we select the 10 topics with the largest number of available questions.
Table~\ref{tab:data_question_topic} presents the complete list of topics per platform.

The number of questions sampled per topic is determined proportionally to the total volume of questions available within that platform.
Within each topic, both questions and corresponding human answers are sampled uniformly at random.
On average, human answers in the test set contain 493.17 characters and 79.80 words.

\subsection{LLM Generation}
\label{subsec:appendix_data_generation}
Stylized responses are generated using the \textbf{Stylized text generation prompt} below, where \texttt{[max\_words]}, \texttt{[title]}, \texttt{[body]}, and \texttt{[style]} are replaced with the maximum word count, question title, question body, and style prompt, respectively.

To control generation prompt length, we truncate question titles to 80 tokens and question bodies to 200 tokens.
We format all inputs using the chat template specific to the LLM used for generation.
For decoding, we set the temperature to 1.0, nucleus sampling threshold ($p$) to 0.95, and maximum token count to 200.
The maximum word count in the prompt is matched with this token count.

\myparagraph{Stylized text generation prompt.}
\texttt{\\
\{\\
"system": "Your task is to answer a question following the provided style instructions. Your response should clearly show the instructed writing style, but should not directly mention any information about the style instruction. Do not explain your answer. Your answer should be within [max\_words] words.",\\
"user": "Answer the following question while adhering to the provided writing style instructions.\textbackslash n\textbackslash n\# Question:\textbackslash n[title][body]\textbackslash n\textbackslash n\# Style Instructions:\textbackslash n[style]"\\
\}
}

\section{Supplementary Material for Section~\ref{sec:method}}
\label{sec:appendix_method}
This section provides additional details on our proposed method.

\subsection{Prompt for Decoding}
\label{subsec:appendix_method_prompt}
We employ the \textbf{Style decoding prompt} below, where \texttt{[max\_words]} is replaced with the maximum word count of 100.
We concatenate \texttt{prefix}, \texttt{projected style representation}, and \texttt{suffix}, which is then provided into the frozen LLM as an input.

\myparagraph{Style decoding prompt.}
\texttt{\\
\{\\
"prefix": "Your task is to describe the writing style from the style embedding vectors, in a single sentence, following this structure: `The author [verb] [specific technique/trait], ...' You can include 1 to 10 techniques/traits, but should not repeat. Do not explain your answer. Your answer should be within [max\_words] words.\textbackslash n\textbackslash n\#\#\# Style Embedding Vectors: \textbackslash n",\\
"suffix": "\textbackslash n\textbackslash n\# Description: \textbackslash n"\\
\}
}

\subsection{Training Details}
\label{subsec:appendix_method_training}
All models are implemented with PyTorch-Lightning and the Huggingface Transformer library.
We use the AdamW optimizer~\cite{loshchilov2018decoupled} and the WSD learning rate schedule~\cite{hu2024minicpm}.
Our model training takes 16 hours using 2 A100 GPUs.

\section{Supplementary Material for Section~\ref{sec:exp1}}
\label{sec:appendix_exp1}
We provide prompts that our LLM baselines and LLM-as-judge evaluations use.

\subsection{LLM baseline prompts}
\label{subsec:appendix_exp1_baseline}
We show three prompts used for three LLM-based description baselines: STYLL~\cite{patel2024styll}, RG~\cite{yang2025registeranalysis}, and the custom prompt we designed.
For RG~\cite{yang2025registeranalysis}, style descriptions were generated in two stages; thus, we present two prompts.
We replace \texttt{[max\_words]} and \texttt{[text]} with the maximum word count and target text, respectively.
The maximum word count is set as 100, matching that of our decoder.

\myparagraph{STYLL.}
\texttt{\\
\{\\
"system": "You are a forensic linguist who knows how to analyze linguistic and stylometric similarites between texts.",\\
"user": "List some adjectives, comma-separated, that describe the writing style of the author of this passage. Strictly output only the style descriptors without any other content.\textbackslash n\textbackslash n\# Passage: \textbackslash n[text]"\\
\}
}

\myparagraph{RG.}
\texttt{\\
\{\\
"system": "You are a forensic linguist who knows how to analyze linguistic and stylometric similarites between texts.",\\
"user": "Analyze the authorship style of this passage in terms of dimensions of register variation according to Douglas Biber.\textbackslash n\textbackslash n\# Passage: \textbackslash n[text]"\\
\}
}

\texttt{\\
\{\\
"system": "You are a forensic linguist who knows how to analyze linguistic and stylometric similarites between texts.",\\
"user": "List some adjectives, comma-separated, that describe the writing style of the author of the target text. Strictly output only the style descriptors without any other content.\textbackslash n\textbackslash n\# Style analysis: \textbackslash n[text]"\\
\}
}

\myparagraph{Custom Prompt.}
\texttt{\\
\{\\
"system": "Your task is to analyze the writing style of the text and describe the style in a single sentence, following this structure: `The author [verb] [specific technique/trait], ...' You can include 1 to 10 techniques/traits, but should not repeat. Do not explain your answer. Your answer should be within [max\_words] words.",\\
"user": "Describe the writing style of the text.\textbackslash n\textbackslash n\# Text: \textbackslash n[text]"\\
\}
}

\subsection{LLM-as-judge evaluation prompt}
\label{subsec:appendix_exp1_evaluation}
We instruct the evaluating LLM with the \textbf{LLM-as-judge score evaluation prompt} below to compute the LLM-as-judge score.
We replace \texttt{[pred]} and \texttt{[ref]} with the predicted and reference styles, respectively.

\myparagraph{LLM-as-judge score evaluation prompt.}
\texttt{\\
You are an expert judge for text similarity. Given the following two passages, rate their semantic similarity on a scale from 0 (completely unrelated) to 10 (nearly identical in meaning).\\
Do not explain your answer. Provide the numeric score only.\\\\
Text A: '[pred]'\\
Text B: '[ref]'\\\\
Similarity score (0–10): 
}


\section{Supplementary Material for Section~\ref{sec:exp2}}
\label{sec:appendix_exp2}
We provide prompts that our baselines use and additional evaluation using different style representation models.

\subsection{LLM baseline prompts}
\label{subsec:appendix_exp2_baseline}
We show four prompts used for four LLM-based style imitation baselines: \citet{bhandarkar2024emulate}, \citet{wang2025imitate}, \citet{jangra2025sptg}, and the custom prompt we designed.
We replace \texttt{[max\_words]}, \texttt{[title]}, \texttt{[body]}, and \texttt{[reference]} with the maximum word count, question title, question body, and the target text, respectively.
The maximum word count is set as 200, matching that of stylized text generation.

\myparagraph{\citet{bhandarkar2024emulate}.}
\texttt{\\
\{\\
"system": "You are an emulator designed to replicate the writing style of a human author.",\\
"user": "\# Task\textbackslash nYour task is to answer the following question while seamlessly integrating with the provided human-authored snippet. Strive to make the answer stylistically indistinguishable from the human-authored text.\textbackslash n\textbackslash n\# Instrucions\textbackslash nThe goal of this task is to mimic the author’s writing style while paying meticulous attention to lexical richness and diversity, sentence structure, punctuation style, special character style, expressions and idioms, overall tone, emotion and mood, or any other relevant aspect of writing style established by the author. Your answer should be within [max\_words] words.\textbackslash n\textbackslash n\# Output Indicator\textbackslash nAs output, exclusively return the text completion without any accompanying explanations or comments.\textbackslash n\textbackslash n\# Question:\textbackslash n[title][body]\textbackslash n\textbackslash n\# Human-authored Text:\textbackslash n[reference]"\\
\}
}

\myparagraph{\citet{wang2025imitate}.}
\texttt{\\
\{\\
"system": "You will be given one or more writing samples from a specific author. Your task is to analyze the author’s style, tone, and voice, then craft an answer that closely mimics their writing based on a provided summary. Your writing should be around [max\_words] words.",\\
"user": "\#\#\# Author's Writing Sample\textbackslash n\textbackslash n[reference]\textbackslash n\textbackslash n\#\#\# Question:\textbackslash n[title][body]\textbackslash n\textbackslash n\#\#\# Instructions\textbackslash n\textbackslash n- Ensure your writing faithfully replicates the author’s style, including tone, word choices, and sentence structure, etc. - Maintain consistency with the author’s voice while accurately reflecting the details of the given summary. - Strive to make your writing indistinguishable from the original author’s work. - Do not output anything other than the writing."\\
\}
}

\myparagraph{\citet{jangra2025sptg}.}
\texttt{\\
\{\\
"system": "You are a writing assistant. Your goal is to address to a user's query to write a text instance based on their preferences.\textbackslash nThe input would comprise of the following elements enclosed in |begin INPUT|...|end INPUT| \textbackslash n - |begin USER\_QUERY|...|end USER\_QUERY| - the user query containing the writing task description and instructions on how to generate the OUTPUT.\textbackslash n - |begin STYLE\_EXAMPLES|...|end STYLE\_EXAMPLES| - contains the written examples that should be used for writing style, tone and voice inspiration.\textbackslash n - Generate the response in |begin OUTPUT|...|end OUTPUT|.\textbackslash n - Depending on the information in input generate the response accordingly.\textbackslash n - Write the response based on instructions in the USER\_QUERY while taking stylistic inspirations from STYLE\_EXAMPLES. Responding with generic response when STYLE\_EXAMPLES are present is undesirable, and therefore you should try your best to incorporate the stylistic features while not leaking any information from STYLE\_EXAMPLES into OUTPUT.",\\
"user": "|begin INPUT|\textbackslash n |begin STYLE\_EXAMPLES|[reference]\textbackslash n |end STYLE\_EXAMPLES|\textbackslash n |begin USER\_QUERY| Answer the following question while mimicking the writing style of the provided reference text. Make sure to not generate infactual information that is not present in the INPUT like dates, names, etc., and instead generate placeholders like [DATE], [NAME], etc. Your answer should be within [max\_words] words.\textbackslash n\textbackslash n\# Question:\textbackslash n[title][body]\textbackslash n|end USER\_QUERY|\textbackslash n|end INPUT|"\\
\}
}

\myparagraph{Custom Prompt.}
\texttt{\\
\{\\
"system": "Your task is to answer a question mimicking the writing style of the reference text. Your response should clearly show the writing style that matches that of the reference text. Do not explain your answer. Your answer should be within [max\_words] words.",\\
"user": "Answer the following question while mimicking the writing style of the provided reference text.\textbackslash n\textbackslash n\# Question:\textbackslash n[title][body]\textbackslash n\textbackslash n\# Reference Text:\textbackslash n[reference]"\\
\}
}

\subsection{Evaluation with other style representations}
\label{subsec:appendix_exp2_additional}
\begin{figure}[t!]
    \centering
    \begin{subfigure}[b]{\columnwidth}
        \centering
        \includegraphics[width=0.9\columnwidth]{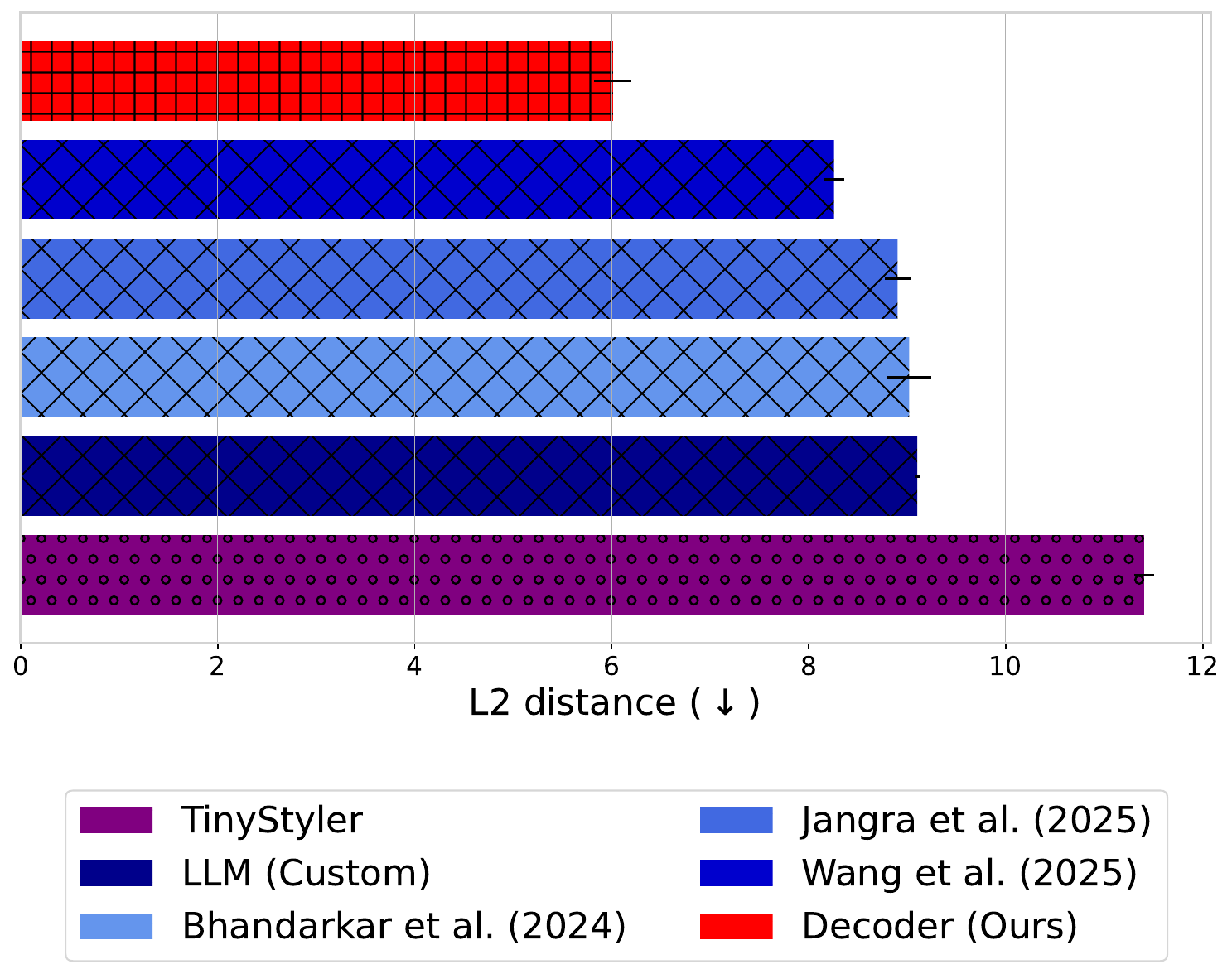}
        \caption{LUAR}
        \label{fig:exp2_luar}
    \end{subfigure}
    \begin{subfigure}[b]{\columnwidth}
        \centering
        \includegraphics[width=0.9\columnwidth]{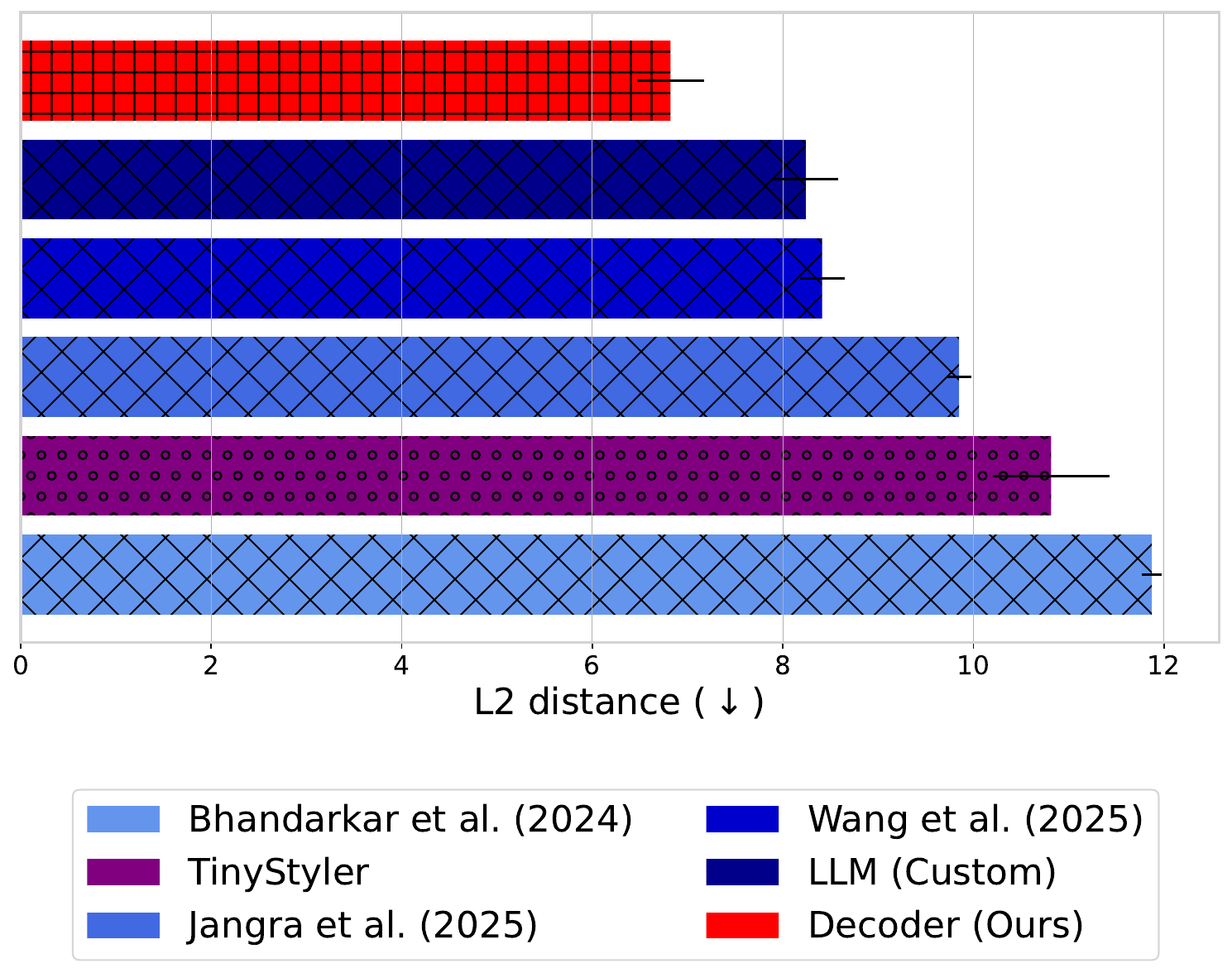}
        \caption{StyleDistance}
        \label{fig:exp2_styledistance}
    \end{subfigure}
    \caption{Our approach yields the smallest L2 distance, measured by other style representations too.}
    \label{fig:exp2_additional_results}
\end{figure}

As additional robustness tests, we reproduce our experiments using two alternative style representation models. Figure~\ref{fig:exp2_additional_results} shows the evaluation in Figure~\ref{fig:exp2_results} with LUAR~\cite{riverasoto2021luar} and StyleDistance~\cite{patel2025styledistance} style embeddings for documents.
Table~\ref{tab:exp2} presents all numeric values for Figures~\ref{fig:exp2_results} and \ref{fig:exp2_additional_results}.

The trends across different representation models agree.
Since our dataset construction and decoder never use these two additional style representation models, there is no potential circularity from these representations.
These results demonstrate that the strong performance of our decoder is not limited to the style representation used to train it.

\begin{table}[t!]
\centering
\resizebox{\columnwidth}{!}{
    \begin{tabular}{llll}
    \textbf{Method} & \textbf{Our Embedding} & \textbf{LUAR} & \textbf{StyleDistance} \\
    \hline
    Decoder (Ours) & 26.07 & 6.01 & 6.82 \\
    LLM (Custom) & 35.39 & 9.10 & 8.24 \\
    Wang et al. (2025) & 73.21 & 8.26 & 8.41 \\
    Jangra et al. (2025) & 100.10 & 8.90 & 9.85 \\
    Bhandarkar et al. (2024) & 102.89 & 9.02 & 11.87 \\
    TinyStyler & 49.97 & 11.40 & 10.82
    \end{tabular}
}
\caption{The numeric values for evaluation in Section~\ref{sec:exp2}.}
\label{tab:exp2}
\end{table}

\section{Supplementary Material for Section~\ref{sec:exp3}}
\label{sec:appendix_exp3}
\begin{figure}[t!]
    \centering
    \begin{subfigure}[b]{\columnwidth}
        \centering
        \includegraphics[width=0.9\columnwidth]{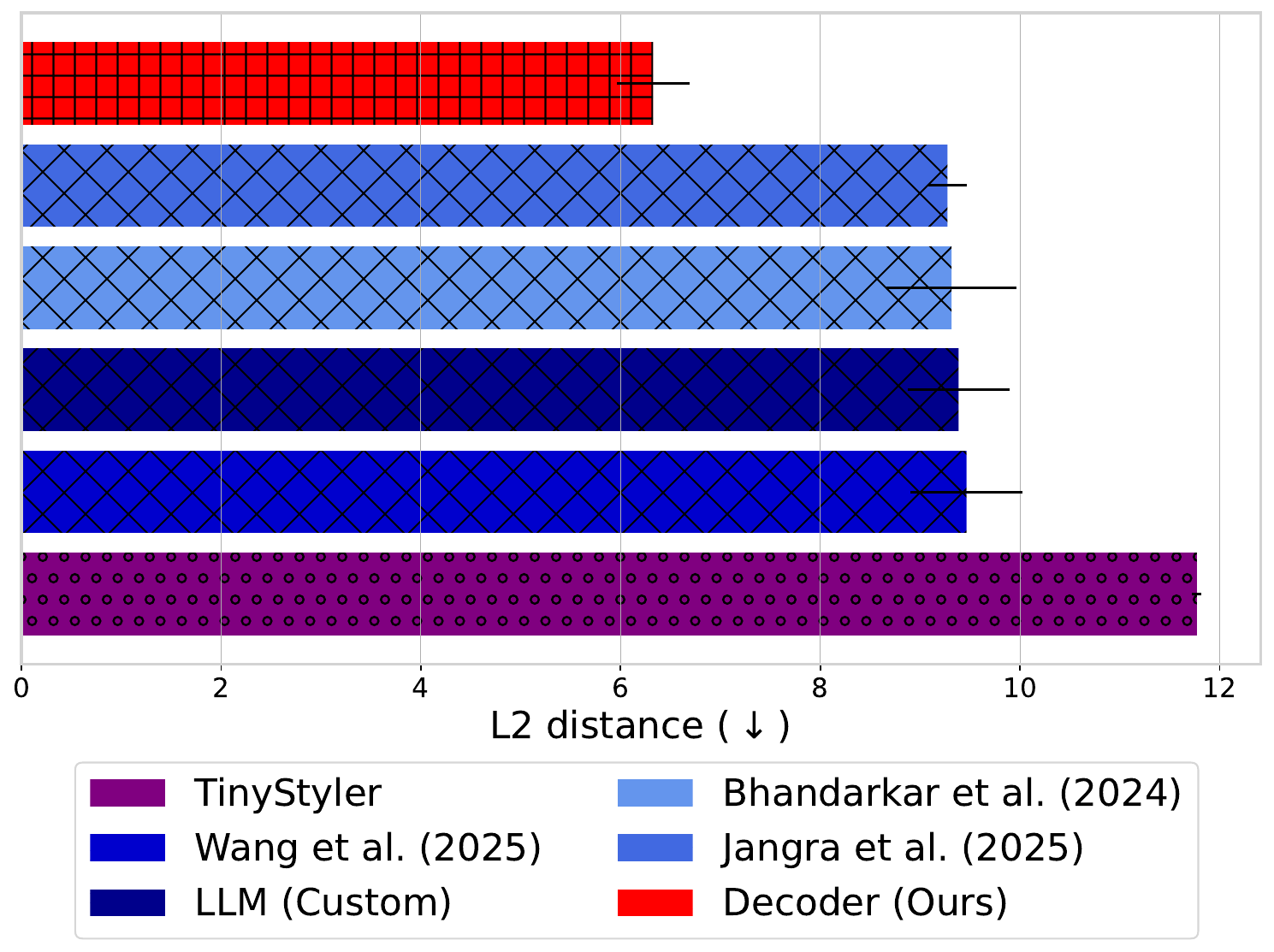}
        \caption{LUAR}
        \label{fig:exp3_luar}
    \end{subfigure}
    \begin{subfigure}[b]{\columnwidth}
        \centering
        \includegraphics[width=0.9\columnwidth]{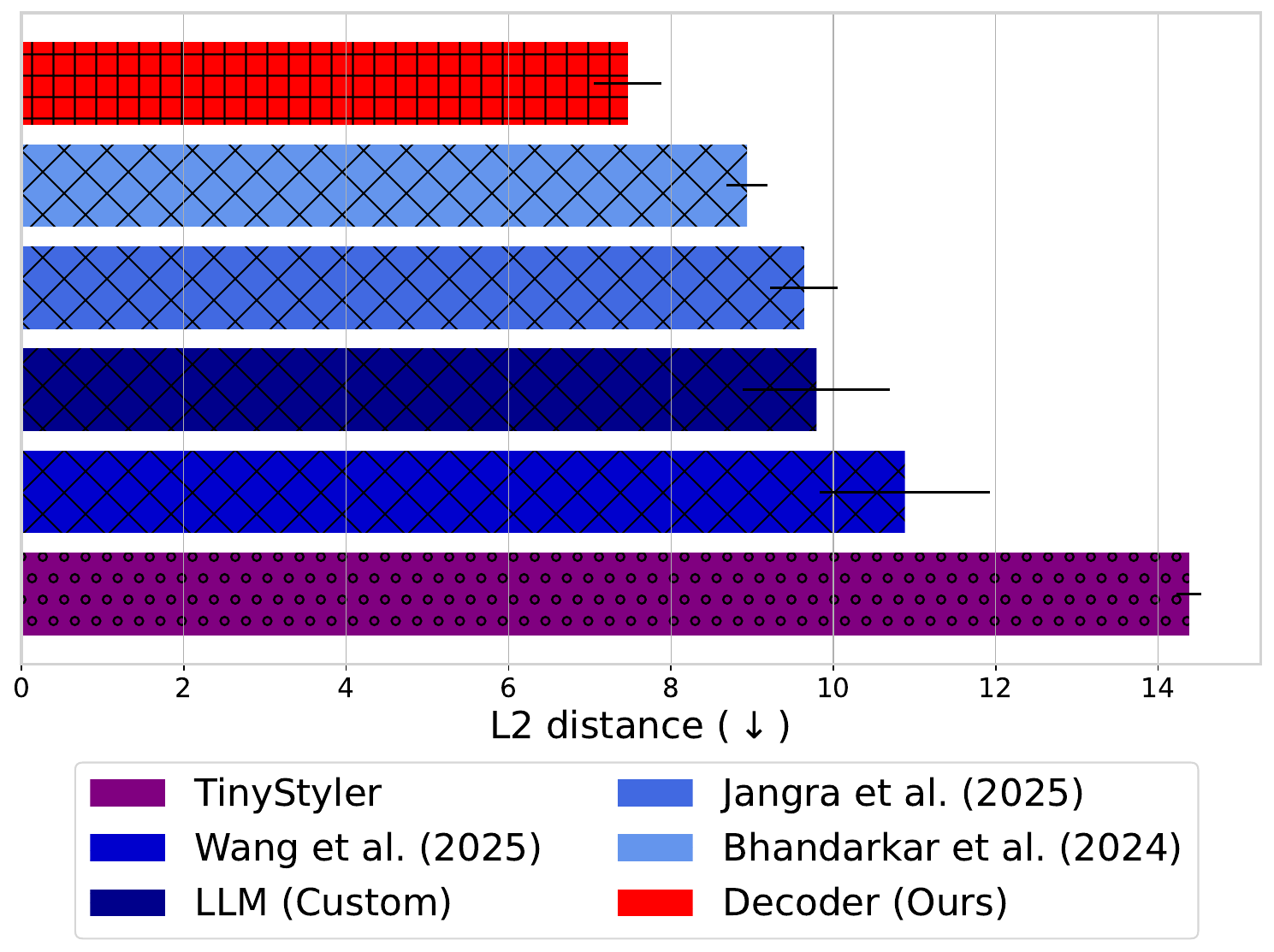}
        \caption{StyleDistance}
        \label{fig:exp3_styledistance}
    \end{subfigure}
    \caption{Our decoder outperforms baselines at generating in human-like styles, measured by other style representations too.}
    \label{fig:exp3_additional_results}
\end{figure}

Figure~\ref{fig:exp3_additional_results} shows the evaluation in Figure~\ref{fig:exp3_results} repeated using two alternative style representation models: LUAR~\cite{riverasoto2021luar} and StyleDistance~\cite{patel2025styledistance}.
Table~\ref{tab:exp3} presents all numeric values for Figures~\ref{fig:exp3_results} and \ref{fig:exp3_additional_results}.

Again, the trends across different representation models align, demonstrating that our decoder is not limited to the style representation used for training.

\begin{table}[t]
\centering
\resizebox{\columnwidth}{!}{
    \begin{tabular}{llll}
    \textbf{Method} & \textbf{Our Embedding} & \textbf{LUAR} & \textbf{StyleDistance} \\
    \hline
    Decoder (Ours) & 27.73 & 6.33 & 7.47 \\
    LLM (Custom) & 37.54 & 9.39 & 9.79 \\
    Wang et al. (2025) & 39.28 & 9.46 & 10.88 \\
    Jangra et al. (2025) & 42.01 & 9.27 & 9.64 \\
    Bhandarkar et al. (2024) & 35.53 & 9.31 & 8.94 \\
    TinyStyler & 54.69 & 11.77 & 14.38
    \end{tabular}
}
\caption{The numeric values for evaluation in Section~\ref{sec:exp3}.}
\label{tab:exp3}
\end{table}

\newpage

\begin{table}[t]
\centering
\resizebox{\columnwidth}{!}{
    \begin{tabular}{lll}
    \textbf{Model} & \textbf{Num. Params.} & \textbf{Huggingface Link} \\
    \hline
    Phi-4 & 14.7B & \href{https://huggingface.co/microsoft/phi-4}{microsoft/phi-4} \\
    Qwen2.5-14B & 14.8B & \href{https://huggingface.co/Qwen/Qwen2.5-14B}{Qwen/Qwen2.5-14B} \\
    OLMo-2-13B & 13.7B & \href{https://huggingface.co/allenai/OLMo-2-1124-13B}{allenai/OLMo-2-1124-13B} \\
    Mistral-Nemo-Instruct-2407 & 12.2B & \href{https://huggingface.co/mistralai/Mistral-Nemo-Instruct-2407}{mistralai/Mistral-Nemo-Instruct-2407} \\
    Ministral-8B-Instruct-2410 & 8.0B & \href{https://huggingface.co/mistralai/Ministral-8B-Instruct-2410}{mistralai/Ministral-8B-Instruct-2410} \\
    Llama-3.1-8B-Instruct & 8.0B & \href{https://huggingface.co/meta-llama/Llama-3.1-8B-Instruct}{meta-llama/Llama-3.1-8B-Instruct} \\
    \end{tabular}
}
\caption{The LLMs used in our study.}
\label{tab:llms}
\end{table}

\begin{table}[t]
\centering
\resizebox{\columnwidth}{!}{
    \begin{tabular}{lr}
    \textbf{Category} & \textbf{\# features} \\
    \hline
    Writing Goal and Intent  & 40 \\
    Tone and Mood & 40 \\
    Readability Level & 40 \\
    Formality & 40 \\
    Narrative Perspective and Voice & 40 \\
    Social and Interpersonal & 40 \\
    Audience Engagement and Interaction & 40 \\
    Emotional Intensity & 38 \\
    Descriptive Density & 40 \\
    Information Density & 40 \\
    Logical Structure and Flow & 39 \\
    Creativity and Typicality & 39 \\
    Abstraction Level & 39 \\
    Temporal Focus & 39 \\
    Technical and Domain Specificity & 40 \\
    Cultural and Regional Influences & 39 \\
    Figurative Language Usage & 38 \\
    Rhetorical Device Usage & 38 \\
    Intertextuality and Allusion & 38 \\
    Sentence Structure and Syntax & 38 \\
    Visual Formatting and Layout & 39 \\
    Function Word Usage & 38 \\
    Word and Expression Choice & 39 \\
    Punctuation Usage & 39 \\
    Special Character and Capitalization Usage & 31 \\
    Acronym and Abbreviation Usage & 39 \\
    \hline
    Total & 1{,}010 \\
    \end{tabular}
}
\caption{The full list of 26 stylistic categories on which we curate style features.}
\label{tab:data_style_category}
\end{table}

\begin{table}[t]
\centering
    \begin{tabular}{ll}
    \textbf{Topic} & \textbf{Platform} \\
    \hline
    Advice  & Reddit \\
    AmItheAsshole  & Reddit \\
    AskMen  & Reddit \\
    AskReddit  & Reddit \\
    askscience  & Reddit \\
    AskWomen  & Reddit \\
    explainlikeimfive  & Reddit \\
    NoStupidQuestions  & Reddit \\
    relationship\_advice  & Reddit \\
    relationships  & Reddit \\
    \hline
    academia  & StackExchange \\
    cooking  & StackExchange \\
    diy  & StackExchange \\
    history  & StackExchange \\
    law  & StackExchange \\
    money  & StackExchange \\
    philosophy  & StackExchange \\
    politics  & StackExchange \\
    scifi  & StackExchange \\
    workplace  & StackExchange \\
    \hline
    business and finance  & Yahoo Answers \\
    computers and internet  & Yahoo Answers \\
    education and reference  & Yahoo Answers \\
    entertainment and music  & Yahoo Answers \\
    family and relationships  & Yahoo Answers \\
    health  & Yahoo Answers \\
    politics and government  & Yahoo Answers \\
    science and mathematics  & Yahoo Answers \\
    society and culture  & Yahoo Answers \\
    sports & Yahoo Answers \\
    \end{tabular}
\caption{The full list of 30 topics from which we source our questions.}
\label{tab:data_question_topic}
\end{table}

\end{document}